# News Recommendation with Attention Mechanism

LIU, Tianrui [1*]  XU, Changxin [2]  QIAO, Yuxin [3]  JIANG, Chufeng [4]  CHEN, Weisheng [5]

[1] University of California San Diego, USA

[2] Northern Arizona University, USA

[3] Universidad Internacional Isabel I de Castilla, Spain

[4] The University of Texas at Austin, USA

[5] Xinhua College of Sun Yat-sen University, China

*LIU, Tianrui is the corresponding author, E-mail: tianrui.liu.ml@gmail.com*

**Abstract:** This paper explores the area of news recommendation, a key component of online information sharing. Initially, we provide a clear introduction to news recommendation, defining the core problem and summarizing current methods and notable recent algorithms. We then present our work on implementing the NRAM (News Recommendation with Attention Mechanism), an attention-based approach for news recommendation, and assess its effectiveness. Our evaluation shows that NRAM has the potential to significantly improve how news content is personalized for users on digital news platforms.

**DOI:** https://doi.org/10.5281/zenodo.10635481

# 1 Introduction

Personalized news recommendation is important for users to find interesting news from massive information. As a heated topic that has wide applications in the industry, it has been extensively studied over decades and has made huge progress. In this paper, we discuss the topic of news recommendation. We implement an attention based model and demonstrate the great performance-boosting modern deep learning techniques bring in.

The following chapter will be structured as follow: in chapter 2, we will briefly introduce the scenario of news recommendation. In Chapter 3, we discuss the details of the implementation of our model. In Chapter 4, we introduce the dataset we use. Chapter 5 will be a recap and conclusion section.

# 2 Problem Formulation

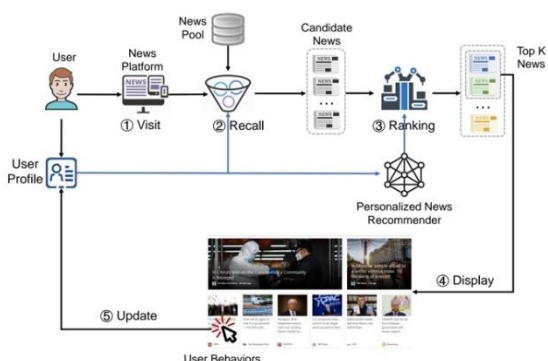

**Figure.1. Scenario of news recommendations**

## 2.1 Workflow of news recommender systems

An example workflow for a personalized news recommender system is depicted in Figure 1. Upon a user's visit to the news platform, the system rapidly recalls a subset of potential articles from a vast news pool, typically selected using rule-based methods. The recommender then ranks these articles in alignment with user interests, which are inferred from the user's profile. Subsequently, the top-ranked news articles—the number of which is denoted by K—are presented to the user. User interactions with these articles are tracked to refine the user profile, thereby enhancing the accuracy of future recommendations.

The described workflow indicates that news recommendation systems must address several challenges. Firstly, the transient nature of news content necessitates an effective strategy for managing the rapid appearance and obsolescence of articles, particularly addressing the cold-start problem. Secondly, the rich textual content of news requires the deployment of sophisticated natural language processing techniques to understand the material and match it to the user's interests and profile. Thirdly, in the absence of direct feedback such as reviews or ratings, user preferences must be deduced from implicit signals, including clicks and time spent on articles. Lastly, although not the focus of this paper, the engineering aspects of scaling news recommendation systems present significant challenges. Considerations include the deployment of high query-per-second (QPS) crawlers, efficient management of large datasets (e.g., inverted indices and text embeddings), and the development of interactive human-computer interfaces. Together, these factors contribute to the complex nature of news recommendation.

## 2.2 News recommendation algorithms

A multitude of comprehensive surveys exists on algorithms for news recommendation. As outlined in several research papers, a typical news recommender system encompasses three main components. Firstly, news modeling forms the foundation of the recommendation process, with a central challenge being the understanding of news content and characteristics. Secondly, user modeling is essential for grasping the personal interests of users in news, where accurately inferring user preferences from profiles and behaviors is crucial. Building upon the representations of news and users derived from the aforementioned models, the system then ranks candidate news articles based on policies that consider the relevance to user interests. With advancements in natural language processing, embeddings for textual information are frequently initialized using pretrained embeddings such as GloVe or BERT. Overall, the distinguishing factors among recent recommendation algorithms lie in their approach to modeling user profiles and news content.

## 2.3 News recommendation datasets

The selection of datasets is a critical aspect of developing personalized news recommendation models. In our study, we utilize the MIND dataset, a benchmark developed by Microsoft and widely recognized in academic research. MIND is comprised of user click logs from Microsoft News, featuring over 1 million users and more than 160,000 English news articles. Each article includes comprehensive textual content such as titles, abstracts, and body text.

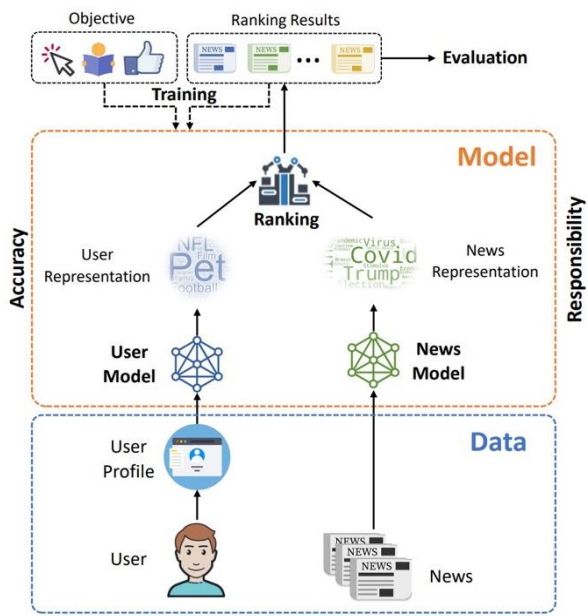

**Figure.2. Architecture of news recommendation models**

## 3 News Recommendation with Attention

The attention mechanism, frequently featured in recent news recommendation literature, originates from the intent to capture word interactions within news text, such as titles and content. Consider the sentence, "a news report is due next Wednesday": terms like "report" and "due" hold substantial informational weight, likely overshadowing less informative words such as "is." Functioning as an effective token mixer, various attention modules can be employed to implicitly learn these connections, thereby enhancing performance in classification tasks.

An attention function is typically defined as mapping a query and a set of key-value pairs to an output, with the query, keys, values, and output all represented as vectors. The output is calculated as a weighted sum of the values. The weight assigned to each value is derived from a compatibility function that assesses the query's relationship with the corresponding key, as illustrated in Figure 3.

This section provides a concise overview of the attention mechanisms implemented in our model, specifically self-attention and additive attention.

### 3.1 Attention Mechanism

#### 3.1.1 Self Attention

In the architecture of a self-attention layer, keys, values, and queries all originate from the same source—specifically, the output of the preceding layer within the encoder. This configuration enables each position within the encoder to consider all positions from the encoder's prior layer. In a parallel fashion, self-attention layers within the decoder permit each position to consider all previous positions, up to the current one. This is essential to maintain the decoder's auto-regressive nature, which requires preventing the flow of information from future positions.

To ensure this property, our implementation of scaled dot-product attention incorporates a masking technique. This approach involves assigning a value of negative infinity (-∞) to all entries in the softmax input that correspond to forbidden connections, effectively nullifying them.

The mechanism of self-attention provides the model with the capability to simultaneously process information from various representation subspaces at disparate positions. However, the use of a singular attention head could limit this ability as it tends to average the attention and may dilute the individual signals.

$$MultiHead(Q, K, V) = Concat(head_1, \ldots, head_h)W^O$$

$$where\ head_i = Attention(QW_i^Q, KW_i^K, VW_i^V)$$

The projections are parameter matrices are of shape

$$W_i^Q \in R^{d_{model} \times d_k}, W_i^K \in R^{d_{model} \times d_k}, W_i^V \in R^{d_{model} \times d_v}$$

And $W_i^O \in R^{dh_{model} \times d_k}$

In this work, we employ h = 15 parallel attention layers, or heads. Due to the reduced dimension of each head, the total computational cost is similar to that of single-head attention with full dimensionality

### 3.1.2 Additive Attention

Additive Attention, also known as Bahdanau Attention, uses a one-hidden layer feed-forward network to calculate the attention alignment score:

$$f_{att}(h_i, s_j) = v_a^T \tanh(W_a[h_i; s_j])$$

Where $v_a$ and $W_a$ are learned attention parameters. Here $h$ refers to the hidden states for the encoder, and $s$ is the hidden states for the decoder. The function above is thus a type of alignment score function. We can use a matrix of alignment scores to show the correlation between source and target words.

Within a neural network, once we have the alignment scores, we calculate the final scores using a softmax function of these alignment scores.

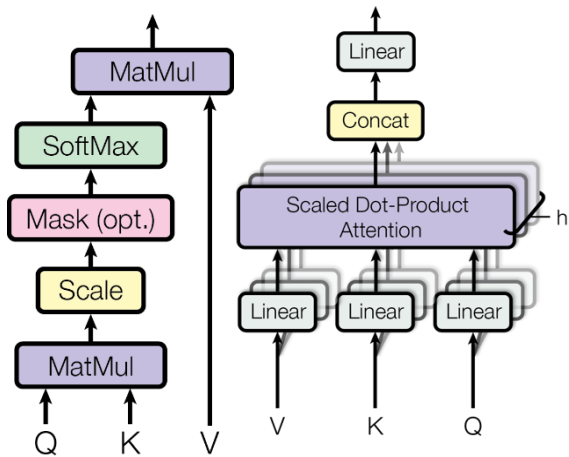

**Figure.3. (left) Scaled Dot-Product Attention. (right) Multi-Head Attention**

## 3.2 News Recommendation with Multi-Head Self-Attention

We proposed NRAM model to cope news recommendation problems with multi-head self attention.

In brief, NMRS models both user and news with multi-head self-attention followed with an additive attention module as aggregator. After that we calculate the dot product of user and news representation to obtain click probability. The training process will be end-to-end.

## 3.3 User Encoder and News Encoder

The news encoder module is used to learn news representations from news titles. It contains three layers. The first one is word embedding, which is used to convert a news title from a sequence of words into a sequence of low-dimensional embedding vectors. Denote a news title with M words. Through this layer it is converted into a vector sequence [$e1, e2, ..., eM$]. This can be conducted using Glove or Bert pre-trained embedding. The second and third layer use token-level multi-head attention network and an additive attention layer to aggregate towards a vector representation perservering information in both token level and semantic level.

The user encoder has similar structure, where we use the news browsed by a user as interaction features. We feed those visited news into a user encoder.

## 3.4 Click Prediction

The user encoder has similar structure, where we use the news browsed by a user as interaction features. We feed those visited news into a user The click predictor module is used to predict the probability of a user browsing a candidate news based on their representations. Denote the representation of a candidate news $Dc$ as r$c$ and the representation of user $u$ as u. The click probability score ˆ$y$ is calculated by the inner product of the representation vectors of user $u$ and the candidate news $D^c$, i.e., ˆ$y = u^T r_c$. We also explored other probability computation methods such as multi-layer neural networks. We find that the inner product is not only the one with the best time efficiency but also the one with the best performance.

## 3.5 Training and Measurement

Noticing that the training datasets only contain positive samples, we use negative sampling to supplement the dataset and speed-up the training. For each news browsed by a user which is regarded as a positive sample, we randomly sample $K$ news which are presented in the same session but are not clicked by this user as negative samples. We then jointly predict the click probability scores of the positive news ˆ$y$+ and the $K$ negative news. In this way, we formulate the news click prediction problem as a pseudo $K + 1$-way classification task. We normalize these click probability scores using softmax to compute the posterior click probability of a positive sample.

# 4 Experiment

## 4.1 Data Exploration

In this part we explore the MIND dataset and introduce the method we use to generate the training data.

### 4.1.1 Behavior Data

The behavior data is extracted from user logs. Each line in the behavior data represents a user's click behavior on a news article. The fields in the behavior data are: impression_id, user_id, time, history, impression. history contains a series of news article IDs the user has clicked before the current impression. impression contains a series of news article IDs that were shown to the users, where the postfix "-1" indicates that the user has clicked the news article, and '-0' indicates that the user has not clicked the news article. In general, we consider the user representation is constructed using the news articles he/she browsed in the history.

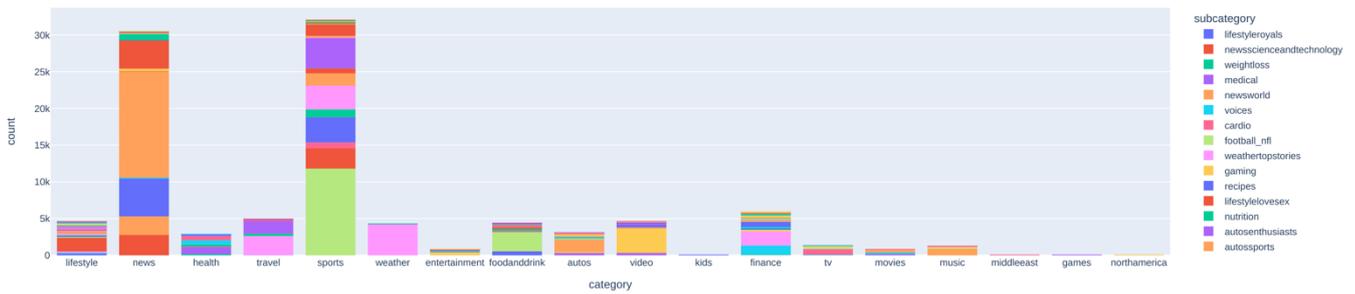

**Figure.4. Category Statistics**

### 4.1.2 News Data

news.tsv contains the news article information. Each line in the news data represents a news article. The fields in the news data are: news_id, category, subcategory, title, abstract, url, title_entities, abstract_entities. title_entities and abstract_entities are the entities in the title and abstract of the news article, respectively.

### 4.1.3 Category Analysis

As shown in Figure 5, we draw the category distribution of the dataset. Besides, each category contains multiple subcategories, which is also shown in the figure below. We can learn from the figure that the dataset contains 18 categories, and news and sports are the dominant ones

### 4.2 Performance Evaluation

We conducted experiments using our model (NRAM) and obtain the following results. Model is implemented in PyTorch. We use early stopping strategy to alleviate the workload. With about two hours of training, we evaluate our model on the testing set and obtain the following results. We also compare our model with another baseline DKN. As we can see, our models with attention (NRAM) outperform previous baselines. (DKN stands for Deep Knowledge-Aware Network for News Recommendation). Which indirectly demonstrates the superiority of the attention mechanism.

| Methods | AUC | MRR | nDCG@5 | nDCG@10 |
|---|---|---|---|---|
| Wide&Deep | 0.6216 | 0.2931 | 0.3908 | 0.3371 |
| DKN | 0.6436 | 0.3028 | 0.3371 | 0.3908 |
| NRAM | **0.6557** | **0.3045** | **0.3389** | **0.4034** |

**Table.1. Experiment results of NRAM and others**

## 5 Future Work And Conclusion

### 5.1 Future work

To achieve better results on the MIND dataset, there are several improvements that are viable.

1. Use a more powerful language model. As we concluded, the attention mechanism is useful in NLP tasks. We could probably replace the self-attention + additive attention model with a Bert encoder. With a stack of attention modules, we can learn a better word/sentence representation (at the cost of larger computation expense).

2. Use more features. As a matter of fact, categorical features like category and subcategory of news can be very helpful. We can encode them using a simple MLP and concatenate these categorical features before feeding into the classifier

### 5.2 Conclusion

In this paper, we reproduce the method NRAM on MIND datasets and demonstrate its availability on the news recommendation task.

In general, we copes both user (sentence level) and news (word level) representation as a series of sequences, and learn the representation with self-attention and additive attention mechanisms. To balance the training process we do negative sampling and transform the problem into a triplet classification problem. NRAM can achieve good results on different metrics, which shows the power of the attention mechanism in handling natural language information.

## Acknowledgments

The authors thank the editor and anonymous reviewers for their helpful comments and valuable suggestions.

## Funding

Not applicable.

# Institutional Review Board Statement

Not applicable.

# Informed Consent Statement

Not applicable.

# Data Availability Statement

The original contributions presented in the study are included in the article/supplementary material, further inquiries can be directed to the corresponding author.

# Conflict of Interest

The authors declare that the research was conducted in the absence of any commercial or financial relationships that could be construed as a potential conflict of interest.

# Publisher's Note

All claims expressed in this article are solely those of the authors and do not necessarily represent those of their affiliated organizations, or those of the publisher, the editors and the reviewers. Any product that may be evaluated in this article, or claim that may be made by its manufacturer, is not guaranteed or endorsed by the publisher.

# Author Contributions

Not applicable.

# About the Authors

**LIU, Tianrui**

Tianrui Liu obtained his Master of Science degree in machine learning and data science from University of California San Diego. His research interests include machine learning, natural language processing, recommendation systems and robotics.

**XU, Changxin**

Affiliation: Northern Arizona University.

**QIAO, Yuxin**

Affiliation: Universidad Internacional Isabel I de Castilla, Spain.

**JIANG, Chufeng**

Affiliation: Department of Computer Science, The University of Texas at Austin.

**CHEN, Weisheng**

Affiliation: Xinhua College of Sun Yat-sen University, China.